# Coherent Keyphrase Extraction via Web Mining

Peter D. Turney
Institute for Information Technology, National Research Council of Canada
M-50 Montreal Road, Ottawa, Ontario, Canada, K1A 0R6
peter.turney@nrc-cnrc.gc.ca

## Abstract

Keyphrases are useful for a variety of purposes, including summarizing, indexing, labeling, categorizing, clustering, highlighting, browsing, and searching. The task of automatic keyphrase extraction is to select keyphrases from within the text of a given document. Automatic keyphrase extraction makes it feasible to generate keyphrases for the huge number of documents that do not have manually assigned keyphrases. A limitation of previous keyphrase extraction algorithms is that the selected keyphrases are occasionally incoherent. That is, the majority of the output keyphrases may fit together well, but there may be a minority that appear to be outliers, with no clear semantic relation to the majority or to each other. This paper presents enhancements to the Kea keyphrase extraction algorithm that are designed to increase the coherence of the extracted keyphrases. The approach is to use the degree of statistical association among candidate keyphrases as evidence that they may be semantically related. The statistical association is measured using web mining. Experiments demonstrate that the enhancements improve the quality of the extracted keyphrases. Furthermore, the enhancements are not domain-specific: the algorithm generalizes well when it is trained on one domain (computer science documents) and tested on another (physics documents).

## 1 Introduction

A journal article is often accompanied by a list of keyphrases, composed of about five to fifteen important words and phrases that express the primary topics and themes of the paper. For an individual document, keyphrases can serve as a highly condensed summary, they can supplement or replace the title as a label for the document, or they can be highlighted within the body of the text, to facilitate speed reading (skimming). For a collection of documents, keyphrases can be used for indexing, categorizing (classifying), clustering, browsing, or searching. Keyphrases are most familiar in the context of journal articles, but many other types of documents could benefit from the use of keyphrases, including web pages, email messages, news reports, magazine articles, and business papers.

The vast majority of documents currently do not have keyphrases. Although the potential benefit is large, it would not be practical to manually assign keyphrases to them. This is the motivation for developing algorithms that can automatically supply keyphrases for a document. Section 2.1 discusses past work on this task.

This paper focuses on one approach to supplying keyphrases, called *keyphrase extraction*. In this approach, a document is decomposed into a set of phrases, each of which is considered as a possible candidate keyphrase. A supervised learning algorithm is taught to classify candidate phrases as *keyphrases* and *non-keyphrases*. The induced classification model is then used to extract keyphrases from any given document [Turney, 1999, 2000; Frank *et al.*, 1999; Witten *et al.*, 1999, 2000].

A limitation of prior keyphrase extraction algorithms is that the output keyphrases are at times incoherent. For example, if ten keyphrases are selected for a given document, eight of them might fit well together, but the remaining two might be outliers, with no apparent semantic connection to the other eight or to each other. Informal analysis of many machine-extracted keyphrases suggests that these outliers almost never correspond to author-assigned keyphrases. Thus discarding the incoherent candidates might improve the quality of the machine-extracted keyphrases.

Section 2.2 examines past work on measuring the coherence of text. The approach used here is to measure the degree of statistical association among the candidate phrases [Church and Hanks, 1989; Church *et al.*, 1991]. The hypothesis is that semantically related phrases will tend to be statistically associated with each other, and that avoiding unrelated phrases will tend to improve the quality of the output keyphrases.

Section 3 describes the Kea keyphrase extraction algorithm [Frank *et al.*, 1999; Witten *et al.*, 1999, 2000]. Each candidate phrase is represented in Kea as a feature vector for classification by a supervised learning algorithm. Four different sets of features are evaluated in this paper. Two of the sets have been used in the past [Frank *et al.*, 1999] and the other two are introduced here, to address the problem of incoherence. The new features are based on the use of web mining to measure the statistical association among the candidate phrases [Turney 2001].

In Section 4, experiments show that the new web mining features significantly improve the quality of the extracted keyphrases, using the author's keyphrases as a benchmark. In the first experiment (Section 4.1), the algorithm is trained and tested on the same domain (computer science documents). In the second experiment (Section 4.2), the algorithm is trained on one domain (computer science) and tested on another (physics). In both cases, the new features result in a significant improvement in performance. In contrast, one of the old feature sets works well in the first case (intradomain) but poorly in the second case (interdomain).

Section 5 considers limitations and future work. Section 6 presents the conclusions.

## 2 Related Work

Section 2.1 discusses related work on generating keyphrases and Section 2.2 considers related work on measuring coherence.

### 2.1 Assignment versus Extraction

There are two general approaches to automatically supplying keyphrases for a document: *keyphrase assignment* and *keyphrase extraction*. Both approaches use supervised machine learning from examples. In both cases, the training examples are documents with manually supplied keyphrases.

In keyphrase assignment, there is a predefined list of keyphrases (in the terminology of library science, a *controlled vocabulary* or *controlled index terms*). These keyphrases are treated as classes, and techniques from text classification (text categorization) are used to learn models for assigning a class to a given document [Leung and Kan, 1997; Dumais *et al.*, 1998]. Usually the learned models will map an input document to several different controlled vocabulary keyphrases.

In keyphrase extraction, keyphrases are selected from within the body of the input document, without a predefined list. When authors assign keyphrases without a controlled vocabulary (in library science, *free text keywords* or *free index terms*), typically from 70% to 90% of their keyphrases appear somewhere in the body of their documents [Turney, 1999]. This suggests the possibility of using author-assigned free text keyphrases to train a keyphrase extraction system. In this approach, a document is treated as a set of candidate phrases and the task is to classify each candidate phrase as either a *keyphrase* or *non-keyphrase* [Turney, 1999, 2000; Frank *et al.*, 1999; Witten *et al.*, 1999, 2000].

Keyphrase extraction systems are trained using a corpus of documents with corresponding free text keyphrases. The GenEx keyphrase extraction system consists of a set of parameterized heuristic rules that are tuned to the training corpus by a genetic algorithm [Turney, 1999, 2000]. The Kea keyphrase extraction system uses a naïve Bayes learning method to induce a probabilistic model from the training corpus [Frank *et al.*, 1999; Witten *et al.*, 1999, 2000]. After training, both systems allow the user to specify the desired number of keyphrases to be extracted from a given input document.

In experimental evaluations, using independent testing corpora, GenEx and Kea achieve roughly the same level of performance, measured by the average number of matches between the author-assigned keyphrases and the machine-extracted keyphrases [Frank *et al.*, 1999; Witten *et al.*, 1999, 2000]. Depending on the corpus, if the user requests five keyphrases to be output, an average of 0.9 to 1.5 of the output phrases will match with the author-assigned keyphrases [Frank *et al.*, 1999]. If the user requests fifteen keyphrases, an average of 1.8 to 2.8 will match with the author's keyphrases.

These performance numbers are misleadingly low, because the author-assigned keyphrases are usually only a small subset of the set of good quality keyphrases for a given document. A more accurate picture is obtained by asking human readers to rate the quality of the machine's output. In a sample of 205 human readers rating keyphrases for 267 web pages, 62% of the 1,869 phrases output by GenEx were rated as "good", 18% were rated as "bad", and 20% were left unrated [Turney, 2000]. This suggests that about 80% of the phrases are acceptable (not "bad") to human readers, which is sufficient for many applications.

Asking human readers to rate machine-extracted keyphrases is a costly process. It is much more economical to use author-assigned keyphrases as a benchmark for evaluating the machine-extracted keyphrases. Jones and Paynter [2001, 2002] argue that it is reasonable to use comparison with authors as a surrogate for rating by human readers. They show (1) the vast majority of Kea keyphrases are rated positively by human readers, (2) the performance of Kea is comparable to GenEx, according to human ratings, (3) readers generally rate author-assigned keyphrases as good, and (4) different human readers tend to agree on the rating of keyphrases.

Furthermore, Gutwin *et al.* [1999] show that machine-extracted keyphrases are useful in an application. The Keyphind system uses keyphrases extracted by Kea in a new kind of search engine. For certain kinds of browsing tasks, Keyphind was judged to be superior to a traditional search engine [Gutwin *et al.*, 1999]. In summary, there is solid evidence that author-assigned keyphrases are a good benchmark for training and testing keyphrase extraction algorithms, but it should be noted that the performance numbers underestimate the actual quality of the output keyphrases, as judged by human readers.

### 2.2 Coherence

An early study of coherence in text was the work of Halliday and Hasan [1976]. They argued that coherence is created by several devices: the use of semantically related terms, coreference, ellipsis, and conjunctions. The first device, semantic relatedness, is particularly useful for isolated words and phrases, outside of the context of sentences and paragraphs. Halliday and Hasan [1976] called this device *lexical cohesion*. Morris and Hirst [1991] computed lexical cohesion by using a thesaurus to measure the relatedness of words. Recent work on text summarization has used lexical cohesion in an effort to improve the coherence of machine-generated summaries. Barzilay and Elhadad [1997] used the WordNet thesaurus to measure lexical cohesion in their approach to summarization.

Keyphrases are often specialized technical phrases of two or three words that do not appear in a thesaurus such as

WordNet. In this paper, instead of using a thesaurus, statistical word association is used to estimate lexical cohesion. The idea is that phrases that often occur together tend to be semantically related.

There are many statistical measures of word association [Manning and Schütze, 1999]. The measure used here is Pointwise Mutual Information (PMI) [Church and Hanks, 1989; Church *et al.*, 1991]. PMI can be used in conjunction with a web search engine, which enables it to effectively exploit a corpus of about one hundred billion words [Turney, 2001]. Experiments with synonym questions, taken from the Test of English as a Foreign Language (TOEFL), show that word association, measured with PMI and a web search engine, corresponds well to human judgements of synonymy relations between words [Turney, 2001].

## 3 Kea: Four Feature Sets

Kea generates candidate phrases by looking through the input document for any sequence of one, two, or three consecutive words. The consecutive words must not be separated by punctuation and must not begin or end with stop words (such as "the", "of", "to", "and", "he", etc.). Candidate phrases are normalized by converting them to lower case and stemming them. Kea then uses the naïve Bayes algorithm [Domingos and Pazzani, 1997] to learn to classify the candidate phrases as either *keyphrase* or *non-keyphrase*.[1]

In the simplest version of Kea, candidate phrases are classified using only two features: *TF×IDF* and *distance* [Frank *et al.*, 1999; Witten *et al.*, 1999, 2000]. In the following, this is called the *baseline feature set* (Section 3.1). Another version of Kea adds a third feature, keyphrase frequency, yielding the *keyphrase frequency feature set* (Section 3.2) [Frank *et al.*, 1999]. This paper introduces a new set of features for measuring coherence. When combined with the baseline features, the result is the *coherence feature set* (Section 3.3). When combined with the keyphrase frequency feature set, the result is the *merged feature set* (Section 3.4).

After training, given a new input document and a desired number of output phrases, *N*, Kea converts the input document into a set of candidate phrases with associated feature vectors. It uses the naïve Bayes model to calculate the probability that the candidates belong to the class *keyphrase*, and then it outputs the *N* candidates with the highest probabilities.

### 3.1 Baseline Feature Set

*TF×IDF* (Term Frequency times Inverse Document Frequency) is commonly used in information retrieval to assign weights to terms in a document [van Rijsbergen, 1979]. This numerical feature assigns a high value to a phrase that is relatively frequent in the input document (the *TF* component), yet relatively rare in other documents (the

---

[1] The following experiments use Kea version 1.1.4, which is available at http://www.nzdl.org/Kea/.

*IDF* component). There are many ways to calculate *TF×IDF*; see Frank *et al.* [1999] for a description of their method. The *distance* feature is, for a given phrase in a given document, the number of words that precede the first occurrence of the phrase, divided by the number of words in the document. The baseline feature set consists of these two features.

The *TF×IDF* and *distance* features are real-valued. Kea uses Fayyad and Irani's [1993] algorithm to discretize the features. This algorithm uses a Minimum Description Length (MDL) technique to partition the features into intervals, such that the entropy of the class is minimized with respect to the intervals and the information required to specify the intervals.

The naïve Bayes algorithm applies Bayes' formula to calculate the probability of membership in a class, using the ("naïve") assumption that the features are statistically independent. Suppose that a candidate phrase has the feature vector <*T, D*>, where *T* is an interval of the discretized *TF×IDF* feature and *D* is an interval of the discretized *distance* feature. Using Bayes' formula and the independence assumption, we can calculate the probability that the candidate phrase is a keyphrase, p(*key*|*T, D*), as follows [Frank *et al.*, 1999]:

$$\mathrm{p}(key\mid T,D) = \frac{\mathrm{p}(T\mid key)\cdot \mathrm{p}(D\mid key)\cdot \mathrm{p}(key)}{\mathrm{p}(T,D)} \quad (1)$$

The probabilities in this formula are readily estimated by counting frequencies in the training corpus.

### 3.2 Keyphrase Frequency Feature Set

The keyphrase frequency feature set consists of *TF×IDF* and *distance* plus a third feature, *keyphrase-frequency* [Frank *et al.*, 1999]. For a phrase *P* in a document *D* with a training corpus *C*, *keyphrase-frequency* is the number of times *P* occurs as an author-assigned keyphrase for all documents in *C* except *D*. Like the other features, *keyphrase-frequency* is discretized [Fayyad and Irani, 1993]. Equation (1) easily expands to include *keyphrase-frequency* [Frank *et al.*, 1999].

The idea here is that a candidate phrase is more likely to be a keyphrase if other authors have used it as a keyphrase. Frank *et al.* [1999] describe *keyphrase-frequency* as "domain-specific", because intuition suggests it will only work well when the testing corpus is in the same domain (e.g., computer science) as the training corpus. This intuition is confirmed in the following experiments (Section 4).

### 3.3 Coherence Feature Set

The coherence feature set is calculated using a two-pass method. The first pass processes the candidate phrases using the baseline feature set. The second pass uses the top *K* most probable phrases, according to the probability estimates from the first pass, as a standard for evaluating the top *L* most probable phrases (*K* < *L*). In the second pass, for each of the top *L* candidates (including the top *K* candidates), new features are calculated based on the statistical association between the given candidate phrase and the top *K* phrases. The hypothesis is that candidates that are semantically related to one or more of the top *K* phrases will tend to be more coherent, higher quality keyphrases.

The coherence feature set consists of 4+2*K* features. The experiments use $K = 4$ and $L = 100$, thus there are twelve features. The first four features are based on the first pass with the baseline model. The first two features are *TF×IDF* and *distance*, copied exactly from the first pass. The third feature is *baseline_rank*, the rank of the given candidate phrase in the list of the top *L* phrases from the first pass, where the list is sorted in order of estimated probability. The fourth feature, *baseline_probability*, is the estimated probability, p(*key*|*T, D*), according to the baseline model.

The remaining eight features are based on statistical association, calculated using Pointwise Mutual Information (PMI) and a web search engine [Turney, 2001]. The following experiments use the AltaVista search engine. Let hits(*query*) be the number of hits (matching web pages) returned by AltaVista when given a query, *query*. Features five to eight are *rank_low$_1$*, *rank_low$_2$*, *rank_low$_3$*, and *rank_low$_4$*. For $i = 1$ to 100 and $j = 1$ to 4, *rank_low$_j$* is the normalized rank of the candidate phrase, *phrase$_i$*, when sorted by the following score:

$$\text{score\_low}_j(phrase_i) = \frac{\text{hits}(low_i \text{ NEAR } low_j)}{\text{hits}(low_i)} \quad (2)$$

For the query to AltaVista, *phrase$_i$* is converted to lower case, *low$_i$*, but it is not stemmed. The rank of score_low$_j$(*phrase$_i$*) is normalized by converting it to a percentile. The ranking and normalization are performed for each document individually, rather than the corpus as a whole. The query "*low$_i$* NEAR *low$_j$*" causes AltaVista to restrict the search to web pages in which *low$_i$* appears within ten words of *low$_j$*, in either order. Equation (2) can be derived from the formula for PMI [Turney, 2001]. The intent is that, with *rank_low$_1$* for example, if a candidate phrase, *phrase$_i$*, is strongly semantically connected with *phrase$_1$*, then *phrase$_i$* will likely receive a high score from Equation (2), and hence *rank_low$_1$* will tend to be relatively high.

Features nine to twelve are *rank_cap$_1$*, *rank_cap$_2$*, *rank_cap$_3$*, and *rank_cap$_4$*. For $i = 1$ to 100 and $j = 1$ to 4, *rank_cap$_j$* is the normalized rank of the candidate phrase, *phrase$_i$*, when sorted by the following score:

$$\text{score\_cap}_j(phrase_i) = \frac{\text{hits}(cap_i \text{ AND } low_j)}{\text{hits}(cap_i)} \quad (3)$$

For the query to AltaVista, *phrase$_i$* is converted to *cap$_i$*, in which the first character of each word in *phrase$_i$* is capitalized, but the other characters are in lower case, and the words are not stemmed. The query "*cap$_i$* AND *low$_j$*" is intended to find web pages in which *phrase$_i$* appears in a title or heading (where it would likely be capitalized) and *phrase$_j$* appears anywhere in the page (where it would likely be in lower case). The rank of score_cap$_j$(*phrase$_i$*) is normalized by converting it to a percentile. The plan is that *rank_cap$_1$*, for example, will be relatively high when the appearance of *phrase$_i$* in a heading makes it likely that *phrase$_1$* will appear in the body. That is, *phrase$_i$* would be a good phrase to put in the title of a document that discusses *phrase$_1$*.

### 3.4 Merged Feature Set

The merged feature set consists of the coherence feature set plus *keyphrase-frequency*. As with the coherence feature set, there is a two-pass process, but the first pass now uses the three features of the keyphrase frequency feature set, instead of the baseline features. The second pass now yields thirteen features, the twelve features of the coherence feature set plus *keyphrase-frequency*. The third and fourth features of the coherence feature set, *baseline_rank* and *baseline_probability*, become *key_freq_rank* and *key_freq_probability* in the merged feature set.

## 4 Experiments

The first experiment examines the behaviour of the four feature sets when trained and tested on the same domain. The second experiment trains on one domain (computer science) and tests on another (physics).

### 4.1 Intradomain Evaluation

This experiment compares the four feature sets using the setup of Frank *et al.* [1999]. The corpus is composed of computer science papers from the Computer Science Technical Reports (CSTR) collection in New Zealand. The training set consists of 130 documents and the testing set consists of 500 documents. The *keyphrase-frequency* feature is trained on a separate training set of 1,300 documents. All other features (including the coherence features) are trained using only the 130 training documents. The naïve Bayes algorithm allows the features to be trained separately, since the features are assumed to be statistically independent. The experiments of Frank *et al.* [1999] show that the *keyphrase-frequency* feature benefits from more training than the baseline features. Figure 1 plots the performance on the testing corpus.

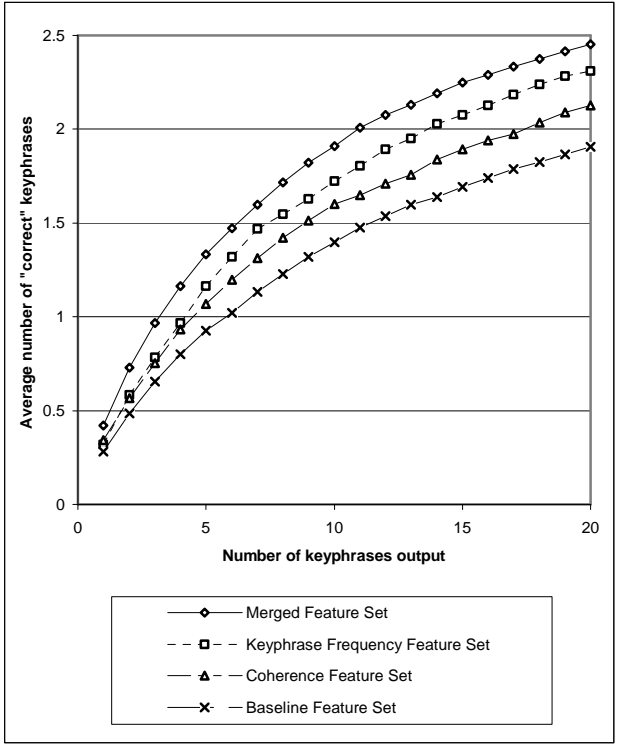
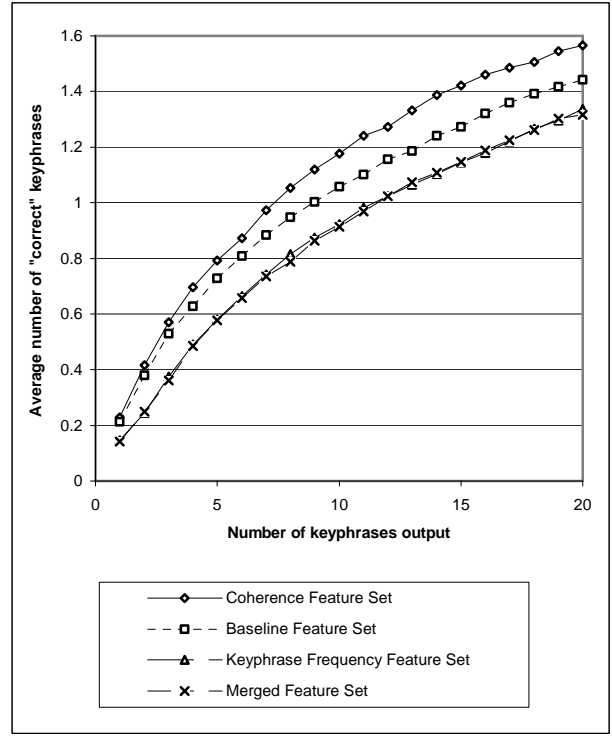

Figure 1. Performance on the CSTR corpus for the four different feature sets. An output keyphrase is considered "correct" when it equals an author-assigned keyphrase, after both are stemmed and converted to lower case.

Figure 2. Performance on the LANL corpus for the four different feature sets. Note that the curves for the keyphrase frequency and merged features sets are intermingled, making it appear that there are only three curves, instead of four.

The paired t-test was used to evaluate the statistical significance of the results [Feelders and Verkooijen, 1995]. There are six possible ways to pair the four curves in Figure 1. The differences between all six pairs are statistically significant with 95% confidence when five or more keyphrases are output. The merged feature set is superior to the keyphrase frequency feature set with 95% confidence when any number of keyphrases are output.

### 4.2 Interdomain Evaluation

This experiment evaluates how well the learned models generalize from one domain to another. The training domain is the CSTR corpus, using exactly the same training setup as in the first experiment. The testing domain consists of 580 physics papers from the arXiv repository at the Los Alamos National Laboratory (LANL). Figure 2 plots the performance on the testing corpus.

The paired t-test shows no significant differences between the keyphrase frequency feature set and the merged feature set. The coherence feature set is superior to the baseline feature set with 95% confidence when four or more keyphrases are output. The baseline feature set is superior to the keyphrase frequency and merged feature sets with 95% confidence when any number of keyphrases are output.

### 4.3 Discussion

Both experiments show that the coherence features are a significant improvement over the baseline features alone. More of the output keyphrases match with the authors' keyphrases, which is evidence that their quality has improved [Jones and Paynter, 2001, 2002]. The second experiment confirms the (previously untested) hypothesis of Frank *et al.* [1999], that the *keyphrase-frequency* feature is domain-specific. This feature actually decreases performance below the baseline in the interdomain evaluation. However, the new coherence features are not domain-specific; they improve over the baseline even when the testing domain is different from the training domain. Furthermore, the coherence features can work synergistically with the *keyphrase-frequency* feature, when the training and testing domains correspond. When they do not correspond, at least the merged feature set is no worse than the keyphrase frequency feature set.

## 5 Limitations and Future Work

The main limitation of the new coherence feature set is the time required to calculate the features. It takes ten queries ($2+2K$, $K = 4$) to AltaVista to calculate one coherence feature vector. Each query takes about one second. At 10 queries per feature vector times 1 second per query times 100 feature vectors per document, we have 1,000 seconds

$((2+2K)L$, $K = 4$, $L = 100$; roughly 15 minutes) per document. The time required by the other aspects of Kea is completely insignificant compared to this. Future work might investigate the use of other, less time-consuming, measures of coherence, such as Morris and Hirst's [1991] measure. However, if we extrapolate current trends in hardware, then it seems that it will not be long before this is no longer an issue, since a local version of AltaVista will easily run on a desktop computer.

# 6 Conclusion

This paper provides evidence that statistical word association can be used to improve the coherence of keyphrase extraction, resulting in higher quality keyphrases, measured by the degree of overlap with the authors' keyphrases. Furthermore, the new coherence features are not domain-specific.

# Acknowledgements

Thanks to Eibe Frank, Gordon Paynter, and the IJCAI reviewers for their helpful comments.